\documentclass[11pt]{article}
\usepackage{eacl2017}
\usepackage{times}
\usepackage{url}
\usepackage{amsmath}
\usepackage{latexsym}
\usepackage[utf8]{inputenc}
\usepackage[T1]{fontenc}
\usepackage{lipsum}
\usepackage{mathabx}
\usepackage{graphicx}
\usepackage{hhline}
\usepackage{caption}%
\usepackage{subcaption}
\usepackage{algorithm}
\usepackage[noend]{algpseudocode}
\usepackage{bm}
\usepackage{multirow} 

\usepackage{xcolor}

\eaclfinalcopy 

\title{On (Commercial) Benefits of Automatic Text Summarization Systems in the News Domain: A Case of Media Monitoring and Media Response Analysis}

\author{Pashutan Modaresi \\ Institute of Computer Science\\ Heinrich Heine University of Düsseldorf\\ {\tt modaresi@cs.uni-duesseldorf.de}
         \AND
         Philipp Gross \\ pressrelations GmbH \\ D-40211 Düsseldorf, Germany \\ {\tt philipp.gross@pressrelations.de} \And
         Siavash Sefidrodi \\ pressrelations GmbH \\ D-40211 Düsseldorf, Germany\\ {\tt siavash.sefidrodi@pressrelations.de}
         \AND
         Mirja Eckhoff \\ pressrelations GmbH \\ D-40211 Düsseldorf, Germany\\{\tt mirja.eckhoff@pressrelations.de} \And
         Stefan Conrad \\ Institute of Computer Science \\ Heinrich Heine University of Düsseldorf\\{\tt conrad@cs.uni-duesseldorf.de}}
         
         \author{Pashutan Modaresi \and  Philipp Gross \and Siavash Sefidrodi \and Mirja Eckhof\\
         pressrelations GmbH \\ D-40211 Düsseldorf, Germany \\ {\tt [first name].[last name]@pressrelations.de}
         \AND
         Stefan Conrad \\ Institute of Computer Science \\ Heinrich Heine University of Düsseldorf\\{\tt conrad@cs.uni-duesseldorf.de}
}

\date{}

\begin{document}
\maketitle
\begin{abstract}
In this work, we present the results of a systematic study to investigate the (commercial) benefits of automatic text summarization systems in a real world scenario. More specifically, we define a use case in the context of media monitoring and media response analysis and claim that even using a simple query-based extractive approach can dramatically save the processing time of the employees without significantly reducing the quality of their work.
\end{abstract}

\section{Introduction}
\label{sec:introduction}

Automatic text summarization has been an evolving field of research. Having started with the pioneering work of Luhn ~\cite{Luhn:1958:ACL:1662353.1662360}, specifically in recent years, automatic text summarization has made remarkable signs of progress with the popularity of deep learning approaches \cite{DBLP:journals/corr/RushCW15,DBLP:conf/naacl/ChopraAR16}.

Providing a formal definition of automatic text summarization is rather a challenging task. This work pursues the following definition: Given a set $Q$ of queries, automatic text summarization is a reductive transformation of a collection of documents $D$ with $|D|> 0$ into a single or multiple target document(s), where the target document(s) are more readable than the documents in $D$ and contain the relevant information of $D$ according to $Q$ \cite{modaresi15}. This definition, comprises both extractive and abstractive approaches, where by extractive we mean methods that select the most salient sentences in a document and by abstractive we mean methods that incorporate language generation to reformulate a document in a reductive way.

Automatic text summarization has been applied to many domains, among which is the news domain the focus of this work. Despite many attempts to improve the performance of summarization systems \cite{Ferreira:2016:ALT:2960811.2967158,Wei:2015:GAM:2766462.2767835}, to the best knowledge of the authors, no systematic study was performed to investigate the (commercial) benefits of the summarization systems in a real world scenario.

We claim that using (even very) simple automatic summarization systems can dramatically improve the workflow of employees without affecting their quality of work. To investigate our claim we define a use case in the context of media monitoring and media response analysis (Section \ref{sec:use_case}) and establish several criteria to measure the effectiveness of the summarization systems in our use case (Section  \ref{sec:evaluation_criteria}). In Section \ref{sec:experiment_setup} we discuss the design of our experiment and report the results in Section \ref{sec:results}. Finally, in Section \ref{sec:conclusions} we conclude our work.

\section{Use Case Definition}
\label{sec:use_case}

We investigate the (commercial) benefits of integrating an automatic summarization system in the semi-automatic workflows of media analysts doing \textit{media monitoring} and \textit{media response analysis} at pressrelations GmbH\footnote{http://www.pressrelations.de/}. In the following, we shortly define the terms as mentioned above.

The goal of media monitoring is to gather all relevant information on specific topics, companies or organizations. To this end, search queries are defined, with which the massive amount of available information can be filtered automatically. Typically, in a post-processing step, the quality of the gathered information is increased using manual filtering by trained media analysts.

In media response analysis, the publications in the media (print media, radio, television, and online media) are evaluated according to various pre-defined criteria. As a result of this, it is possible to deduce whether and how journalists have recorded and processed the PR (Public Relations) messages. Possible questions to be answered in the context of media response analysis are: How are the publications distributed over time? How many listeners, viewers or readers were potentially reached? What are the tonality and opinion tendency of the publications? \cite{grupe2011public}

Typically, analysis results are given to the clients in the form of extensive reports. In the case of textual media, the immense amount of time required to read texts and to write abstracts and reports is a high cost factor in the preparation of media resonance analysis reports. 

We claim that the described process can be partially optimized by incorporating automatic summarization systems, leading to remarkable financial advantages for the companies.

\section{Evaluation Criteria}
\label{sec:evaluation_criteria}

From the commercial and academic point of view, the \textit{quality} of the summaries plays an import role. Various automatic methods such as ROUGE \cite{lin:2004:ACLsummarization}, BLEU \cite{Papineni:2002:BMA:1073083.1073135}, and Pyramid \cite{Nenkova:2007:PMI:1233912.1233913} have been used successfully to evaluate the quality of the summaries. Moreover, manual evaluation has also been incorporated for quality assessment of the summaries \cite{Modaresi:2014:PKU:2684103.2684117}. Another important criterion that is mostly neglected in academic publications is the  \textit{gain in time}, defined as the amount of saved time by a user through the usage of the summaries.

In our use case, the quality of a summary comprises of two aspects: \textit{completeness} and \textit{readability}. The term \textit{completeness}, describes the requirement of a summary to contain all relevant information of an article. The relevance of information is determined based on a \textit{query}. For instance, the query might be a named entity, and we expect that the summary contains all relevant information regarding the named entity.

The term \textit{readability} refers to the coherence and the grammatical correctness of the summary. While the grammatical correctness is defined at the sentence level, the coherence of the summary is determined on the whole text. That means that the sentences of the summary should not only be grammatically correct in isolation, but also they must be coherent to make the summary readable.

Both \textit{completeness} and \textit{readability} are criteria that are difficult to evaluate and define formally, and it has been shown that they are both very subjective criteria, where their assessment varies from person to person \cite{automatic_text_summarization}. In the case of completeness, it is unclear how to formalize the relevance of information, and in in the case of readability the same holds for the concept of coherence.

Therefore, we define the quality of a summary from a practical and commercial point of view. For this, we define the quality of a summary in terms of a binary decision problem where the question to be asked is: \textit{can the produced summary in its current form be delivered to a customer or not?}

Furthermore, in our use case, the \textit{gain in time} is defined as the processing time that can be saved by media analysts, assisted by a summarization system. It should be clear that the reduction of the processing time could lead to the reduction of costs in a company.

In the following section, the design of our experiment with respect to the criteria mentioned above (quality and gain in time) will be explained.
\section{Experiment Setup}
\label{sec:experiment_setup}

To conduct our experiments we incorporated eight media analysts (specialists in writing summaries for customers) and divided them into two equi-sized groups. One group received only the news articles (Group A), and the other one received only the query-based extracted summaries without having access to the original articles. Given a query consisting of a single named entity, both groups were asked to write summaries with the following properties:

\begin{itemize}
    \item The summary should be compact and consist of maximum two sentences.
    \item The summary should contain the main topic of the article and also the most relevant information regarding the query.
\end{itemize}

As previously stated, the summaries created by media analysts were evaluated based on two criteria: \textit{quality} and \textit{gain in time}. The \textit{gain in time} was measured automatically using a web interface by tracking the processing time of the media analysis for creating the text summaries. We interpret the \textit{gain in time} as the answer to the question: \textit{In average, what percentage faster/slower is group A in compare to group B?}. Let $\tilde{t}_A$ and $\tilde{t}_B$ be the average processing times of the media analysts in group A and B respectively. We define \textit{gain in time} as in Equation \ref{eq:gain_in_time}.

\begin{equation}
\label{eq:gain_in_time}
 g(\tilde{t}_A, \tilde{t}_B)  =
\begin{cases}
100 \cdot \left|1 - \frac{\tilde{t}_A}{\tilde{t}_B}\right| & \text{if } \tilde{t}_A \leq \tilde{t}_B\\
100 \cdot \left|1 - \frac{\tilde{t}_B}{\tilde{t}_A}\right| & \text{if } \tilde{t}_A > \tilde{t}_B
\end{cases}
\end{equation}

\noindent
Notice that it holds $g(\tilde{t}_A, \tilde{t}_B) = g(\tilde{t}_B, \tilde{t}_A)$ and $g$ reflects only the magnitude of the saved time and not its direction. The direction can be determined based on the values of $\tilde{t}_A$ and $\tilde{t}_B$.

On the other hand, the quality of the summaries was evaluated by a curator (an experienced media analyst in direct contact with customers). The curator was responsible for evaluating the summaries created by media analysts in both groups and scored them with a zero or a one. With zero meaning that the quality of the summary is not sufficient and the product cannot be delivered to the client and with one meaning the quality of the summary is sufficient enough to be delivered to the customer. Let the vector $q$ of size $m$ be a one-hot vector consisting of 0s and 1s, where the $i$-th element in $q$ represents the evaluation of the curator for the $i$-th summary among the $m$ available summaries. Given that, we compute the average summary quality of a set of summaries by computing the average of its corresponding evaluation vector $q$.

In total, ten news articles were provided to the media analysts. The articles for group A had an average word count of 1438 with the standard deviation being 497. Group B received only the summaries of the articles, created automatically with a heuristic-based approach. The automatically generated summaries had an average length of 81 words with the standard deviation being 23. The pseudocode of the invoked query-based extractive summarizer is depicted in Algorithm \ref{algorithm}.

\begin{algorithm}
\caption{Query-based Summarization}
\begin{algorithmic}[1]
\Procedure{summarize(T, q)}{}
\State $S \gets \emptyset$
\State $T' \gets \text{Segment}(T)$
\State $E \gets \text{EntityDistribution}(T')$
\State $m \gets \text{Median}(E)$ 
\State $E' \gets \emptyset$
\For{\texttt{e in E}}
  \If {freq(e) > m}
      \State $E' \gets E' \cup e$
  \EndIf
\EndFor
\State $S \gets \text{Lead}(T')$
\State $S \gets \text{QueryMatch}(T', Q)$
\State $S \gets \text{CentralEntityMatch}(T', E')$
\State \textbf{return} S
\EndProcedure
\end{algorithmic}
\label{algorithm}
\end{algorithm}

In line 2 the summary $S$ is initialized with an empty set. Given the input text $T$, the text is segmented into sentences and stored in the list $T'$ (line 2). In line 3, the named entities of the text are recognized and stored in a dictionary where each key represent a named entity, and its corresponding value is the frequency of the named entity in the text. Lines 5-9 depict the procedure to select \textit{central named entities}. Let $m$ be the median of the named entities frequencies. A named entity $e$ is called a \textit{central named entity} if its frequency in the text is higher than the twice of the median. In line 10 we add the lead of the news article to the summary, as the lead usually can be interpreted as a  compact summary of the whole article. Afterwards in line 11, the sentences that contain the query $Q$ are added to the summary. Finally, we extend the list of summary sentences with sentences containing the \textit{central named entities} and return the summary.
\section{Results}
\label{sec:results}

In total, we collected 80 summaries created by the media analysts in both groups. For each summary, its processing time and its quality evaluated by a curator was recorded. Based on the collected data, we answered the following questions:

\begin{enumerate}
    \item Intergroup processing time: Is there a significant difference between the processing times of individual media analysts in a group?
    \item Intergroup quality: Is there a significant difference between the quality of the created summaries by the media analysts in a group?
    \item Intragroup processing time: Is there a significant different between the average processing times of media analysts in groups A and B? If so, which group has a faster processing time?
    \item Intragroup quality: Is there a significant difference between the average qualities of created summaries by media analysts in groups A and B? If so, which group created more qualitative summaries?
\end{enumerate}

The remaining of this section reports the answers to the above questions.

\begin{figure*}
\centering
\begin{subfigure}{0.5\textwidth}
  \centering
  \includegraphics[scale=0.45]{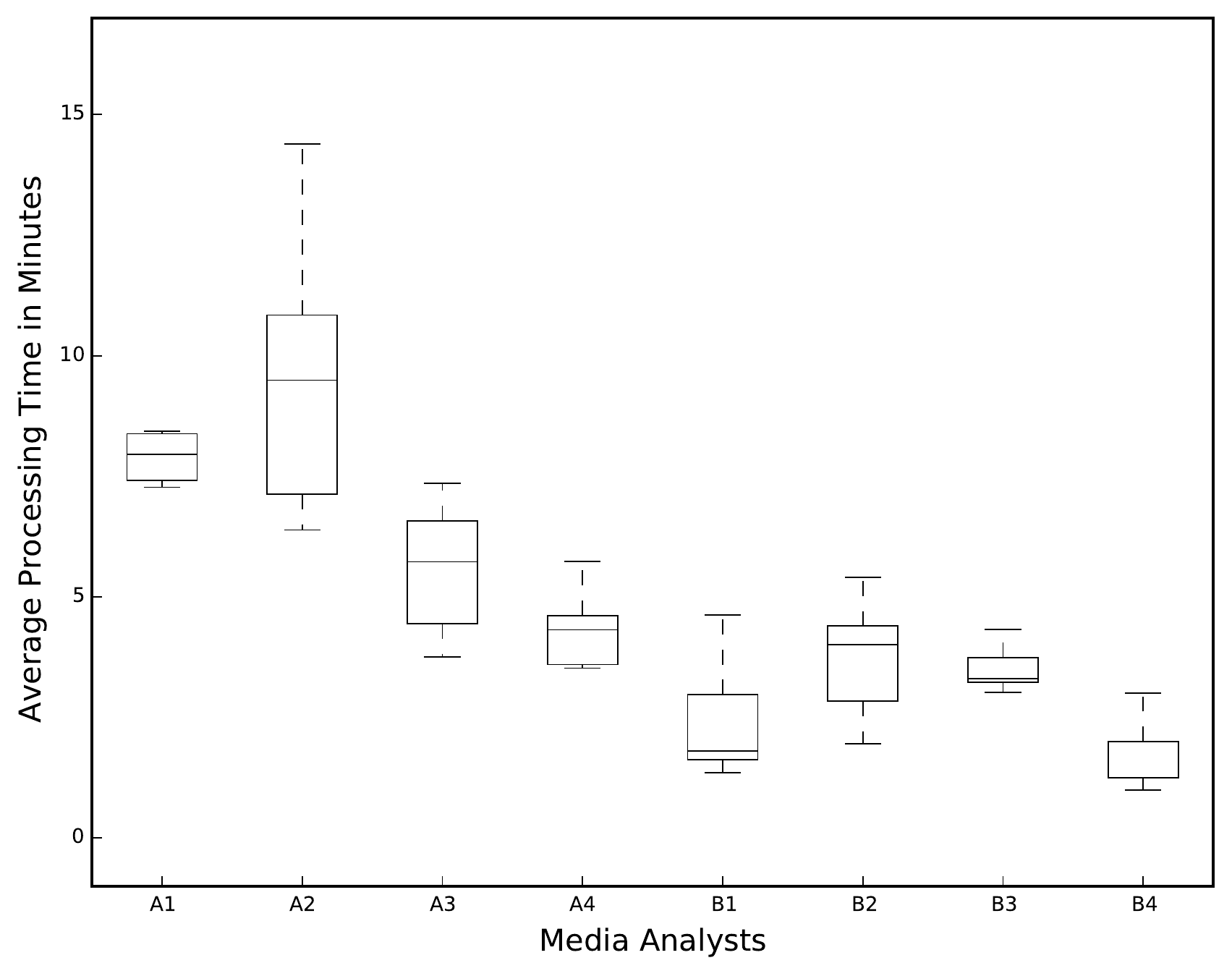}
  \caption{Processing Times of Individual Media Analysts}
  \label{fig:boxplotsA}
\end{subfigure}%
\begin{subfigure}{0.5\textwidth}
  \centering
  \includegraphics[scale=0.45]{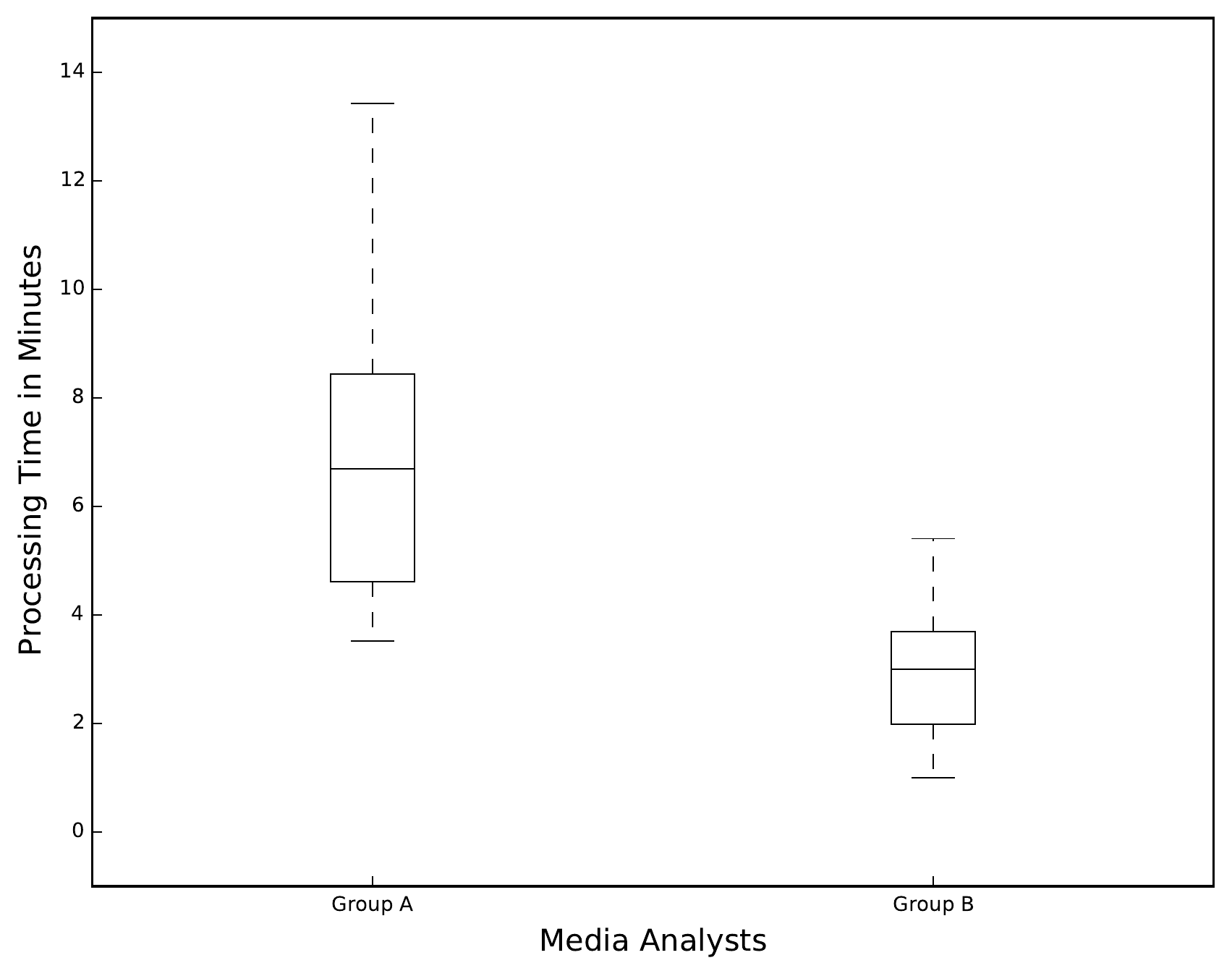}
  \caption{Processing Times of Groups A and B}
  \label{fig:boxplotsB}
\end{subfigure}
\caption{Comparison of the Processing Times}
\label{fig:boxplots}
\end{figure*}

\subsection{Intergroup Processing Time}
\label{sec:intergroup_processing_time}
The processing times of the media analysts in group A (A1-A4) and group B (B1-B4) are visualized using boxplots in Figures \ref{fig:boxplotsA} and \ref{fig:boxplotsB} respectively. In both groups, the differences among the average processing times are observable. Our goal is to investigate whether the differences between the processing times of media analysts is statistically significant.

To compare the means of processing times among the media analysts in a group we use the \textit{one-way analysis of variance} (one-way ANOVA). The null hypothesis in the ANOVA test is that the mean processing times of the media analysts in a group are the same. To perform the ANOVA test we first examine if the requirements of the ANOVA test are satisfied \cite{miller1997beyond}.

The first requirement of the ANOVA test is that the processing times of the individual media analysts are normally distributed. For this, we use the Shapiro-Wilk test \cite{ShapiroWilk1965} with the null hypothesis being that the processing times are normally distributed. Table \ref{tab:shapiro-wilk-processing-time} reports the results of the test.

\begin{table}[ht]
	\def\arraystretch{1.2}
	\small
	\begin{center}
		\begin{tabular}{|c|c|c|}
		\hline
		Media Analyst & $W$ & $p$-value\\
		\hline
		\hline
		A1 & 0.90244& 0.233 \\
		\hline
		A2 & 0.91638 & 0.3277 \\
		\hline
		A3 & 0.76592& 0.0055\\
		\hline
		A4 & 0.73605 & 0.0024 \\
		\hhline{|=|=|=|}
		B1 & 0.85143& 0.0604\\
		\hline
		B2 & 0.95625&0.7425 \\
		\hline
		B3 &0.94609 &0.6226 \\
		\hline
		B4 & 0.93536& 0.5026\\
		\hline
		\end{tabular}
		
	\end{center}
\caption{Shapiro-Wilk Test for Processing Times}\label{tab:shapiro-wilk-processing-time}
\end{table}

In Table \ref{tab:shapiro-wilk-processing-time}, $W$ is the test statistic and we reject the null hypothesis if the $p$-value is less than the chosen significance level $\alpha=0.05$. Thus the null hypothesis will be rejected for A3, A4, and B1, meaning that the processing times of them are not normally distributed. For other media analysts, the normality assumption holds. Although in several cases the normality requirement of the ANOVA test is violated, it is still possible to use the ANOVA test, as it was shown that the ANOVA test is relatively robust to the violation of the normality requirement \cite{kirk2012experimental}.

The second requirement to perform the ANOVA test is that the processing times of the media analysts have equal variances. For this, we use the Bartlett's test \cite{dalgaard2008introductory} with the null hypothesis that the processing times of the media analysts have the same variance. The results of the Bartlett's test for groups A and B are reported in Table \ref{tab:bartlett-processing-time}

\begin{table}[ht]
	\def\arraystretch{1.2}
	\small
	\begin{center}
		\begin{tabular}{|c|c|c|}
		\hline
	    Group & $\chi^2$ & $p$-value\\
		\hline
		\hline
		A & 4.3726& 0.2239 \\
		\hline
		B & 6.9013 & 0.0751 \\
		\hline
		\end{tabular}
		
	\end{center}
\caption{Bartlett's Test for Processing Times}\label{tab:bartlett-processing-time}
\end{table}

In Table \ref{tab:bartlett-processing-time}, $\chi^2$ is the test statistic and we reject the null hypothesis if the $p$-value is less than the chosen significance level $\alpha=0.05$. For both groups, the $p$-value is greater than the significance level, and thus there is no evidence that the variances of processing times of individual media analysts are different.

Having investigated the assumptions of the ANOVA test, we now report the results of the ANOVA test (See Table \ref{tab:anova-processing-time}).

\begin{table}[ht]
	\def\arraystretch{1.2}
	\small
	\begin{center}
		\begin{tabular}{|c|c|c|}
		\hline
	    Group & F value & $p$-value\\
		\hline
		\hline
		A & $1.413 \cdot 10^{33} $& $< 2 \cdot 10^{-16}$ \\
		\hline
		B & $4.2 \cdot 10^{34}$ & $< 2 \cdot 10^{-16}$ \\
		\hline
		\end{tabular}
		
	\end{center}
\caption{ANOVA Test for Processing Times}\label{tab:anova-processing-time}
\end{table}

In Table \ref{tab:anova-processing-time}, the F value is the F test statistic and we reject the null hypothesis if the $p$-value is less than the chosen alpha level $\alpha=0.05$. Thus, the mean processing times of media analysts in group A are not the same and there is a significant difference between them. The same hold for group B.

The so far shown results crystallize an important property of the summarization process. Given the same set of news articles and the same briefing to all media analysts, the average time required by the media analysts within a group to summarize the articles is significantly different from each other. 

\subsection{Intergroup Quality}
The results of the manual evaluation of the summaries by the curator are represented in Table \ref{tab:manual_quality}. In this section, our goal is to systematically investigate whether the qualities of the summaries produced by media analysts in a group are significantly different from each other.

Different from the previous section where we compared the processing times of the media analysts in a group using the ANOVA test, the comparison of the qualities among the media analysts cannot be performed using the ANOVA test (due to the huge violation of the normality assumption). Therefore, we interpret the evaluation results of each media analyst as a Binomial distribution $\operatorname{B} \left({n, p}\right)$ with $n=10$ (number of articles shown to each media analyst) and $p$ being the numbers of times the curator was satisfied with the quality of the summaries created by the media analyst.

\begin{table}[ht]
	\def\arraystretch{1.2}
	\small
	\begin{center}
		\begin{tabular}{cc|c|c|c||c|c|c|c|c|} 
			\cline{2-9}
		    & \multicolumn{4}{ |c|| }{Group A} & \multicolumn{4}{c| }{Group B}  \\ \cline{2-9}
		    
            & \multicolumn{1}{ |c| }{A1} & A2 & A3 & A4 & B1 & B2 & B3 & B4      \\ \cline{1-9}
            
			\multicolumn{1}{ |c| }{Quality} & 0.9  & 1.0 &  0.8  & 0.6  & 0.3 & 0.9 & 1.0 & 0.7   \\ \cline{1-9}
			
			\multicolumn{1}{ |c }{Overall} & \multicolumn{4}{ |c|| }{0.82}  & \multicolumn{4}{ c| }{0.72}   \\ \cline{1-9}
		\end{tabular}
		
	\end{center}
\caption{Results of Manual Evaluation of Quality}\label{tab:manual_quality}
\end{table}

To test whether the qualities of the produced summaries are significantly different from each other, we use the Fisher's Exact Test \cite{dalgaard2008introductory} with the null hypothesis that the qualities are not different from each other. For group A we have a $p$-value of 0.1087, and for group B the $p$-value is 0.0022. Thus the null hypothesis can only be rejected for group B. The so far shown results lead us to the following conclusions: Given the news articles, no significant difference among the qualities of the produced summaries by the media analysts can be observed. Furthermore, given only the automatically created summaries, the media analysts produce summaries with significantly different qualities.

\subsection{Intragroup Processing Time}
So far, we only investigated the intergroup properties. In this section, we answer the question whether there exists a significant difference between the average processing times of group A and B?

In Figure \ref{fig:boxplotsB}, the processing times of the groups A and B are compared using boxplots. Using the Equation \ref{eq:gain_in_time}, we compute the \textit{gain in time} for group B, that is roughly 58\%, meaning that as expected, the media analysts in group B required much less time to create the summaries in compare to the media analysts in group A. Similar to the Section \ref{sec:intergroup_processing_time}, we use the ANOVA test to check the significance of this outcome. The results of the test are reported in Table \ref{tab:anova-intreagroup}.

\begin{table}[ht]
	\def\arraystretch{1.2}
	\small
	\begin{center}
		\begin{tabular}{|c|c|c|}
		\hline
	    Group & F value & $p$-value\\
		\hline
		\hline
		A vs. B & $2.14 \cdot 10^{33} $& $< 2 \cdot 2^{-16}$ \\
		\hline
		\end{tabular}
		
	\end{center}
\caption{ANOVA Test for  Intragroup Processing Times}\label{tab:anova-intreagroup}
\end{table}

In Table \ref{tab:anova-intreagroup}, F value is the F test statistic and we reject the null hypothesis if the $p$-value is less than the chosen alpha level $\alpha=0.05$. Thus, the processing times of media analysts in group B are significantly lower than the processing times of media analysts in group A. 

The results show that using a simple query-based extractive summarization system, the media analysts had a significant gain in time by the process of creating the text summaries.

\subsection{Intragroup Quality}
In the final step, we compare the quality of the produced summaries between both groups and answer the question whether there is a significant difference between the qualities? To answer this question we perform the Fisher's Exact Test and obtain the $p$-value of 0.4225. Thus the null hypothesis of the test cannot be rejected and we conclude that the qualities of the summaries among both groups are not significantly different. 

Using the results above, we conclude that providing the media analysts with automatically created summaries does not have a negative impact on the quality of the summaries they generated and no significant difference in quality could be observed in compare to the media analysts that had access to the full new articles.
\section{Conclusions}
\label{sec:conclusions}

To investigate the (commercial) benefits of the summarization systems, we designed an experiment where two groups of media analysts were given the task to summarize news articles. Group A received the whole news articles and group B received only the automatically created text summaries. In summary, we showed that:
\begin{itemize}
    \item The average time required by the media analysts within a group to summarize the articles is significantly different from each other.
    \item Given the news articles, no significant difference among the qualities of the produced summaries by the media analysts can be observed. Furthermore, given only the automatically created summaries, the media analysts produce summaries with significantly different qualities.
    \item The media analysts had a significant gain in time by the process of creating the text summaries (58\%).
    \item Providing the media analysts with automatically created summaries does not have a negative impact on the quality of the summaries they generated
\end{itemize}

The results mentioned above indicate that incorporating even simple summarization systems can dramatically improve the workflow of the employees.

For future work we plan to repeat our experiment with more sophisticated summarization algorithms and compare the \textit{gain in time} to our baseline setting. Furthermore, we plan to increase the number of media analysts to obtain more reliable results.

\section*{Acknowledgments}
This work was funded by the German Federal Ministry of Economics and Technology under the ZIM program (Grant No. KF2846504).

\bibliography{eacl2017}
\bibliographystyle{eacl2017}

\end{document}